\documentclass{article}

\usepackage{arxiv}

\usepackage[utf8]{inputenc} 
\usepackage[T1]{fontenc}    
\usepackage{hyperref}       
\hypersetup{colorlinks,linkcolor={blue},citecolor={green},urlcolor={red}}  
\usepackage{cleveref}
\usepackage{url}            
\usepackage{booktabs}       
\usepackage{amsfonts}       
\usepackage{nicefrac}       
\usepackage{microtype}      
\usepackage{lipsum}
\usepackage{graphicx}
\usepackage{float}
\graphicspath{ {./figures/} }

\title{Developing novel ligands with enhanced binding affinity for the sphingosine 1-phosphate receptor 1 using machine learning}

\author{
 Colin Zhang\\
  Carlmont High School\\
  Belmont, CA 94002 \\
 \and
 \textbf{Yang Ha}\\
  Lawrence Berkeley National Laboratory\\
  Berkeley, CA 94720 \\
}

\begin{document}
\maketitle
\begin{abstract}
Multiple sclerosis (MS) is a debilitating neurological disease affecting nearly one million people in the United States. Sphingosine-1-phosphate receptor 1, or S1PR1, is a protein target for MS. Siponimod, a ligand of S1PR1, was approved by the FDA in 2019 for MS treatment, but there is a demonstrated need for better therapies. To this end, we finetuned an autoencoder machine learning model that converts chemical formulas into mathematical vectors and generated over 500 molecular variants based on siponimod, out of which 25 compounds had higher predicted binding affinity to S1PR1. The model was able to generate these ligands in just under one hour. Filtering these compounds led to the discovery of six promising candidates with good drug-like properties and ease of synthesis. Furthermore, by analyzing the binding interactions for these ligands, we uncovered several chemical properties that contribute to high binding affinity to S1PR1. This study demonstrates that machine learning can accelerate the drug discovery process and reveal new insights into protein-drug interactions.
\end{abstract}

\section{Introduction}
\label{sec:intro}
Multiple sclerosis (MS) is a chronic disease that harms the central nervous system of the human body, including the brain, spinal cord, and optic nerves. It causes symptoms such as cognitive loss, fatigue, and loss of vision, significantly impacting the quality of life for those who are affected. Scientists have developed effective therapies for MS, with a focus on targeting specific receptors in the immune system \cite{bourque2021current, loma2011multiple}. One such receptor is the sphingosine 1-phosphate receptor 1 (S1PR1), which plays a key role in the immune response and has been the target of several approved therapies \cite{mcginley2021sphingosine}. In 2019, the FDA approved the use of siponimod, a ligand that binds to S1PR1, as a treatment for relapsing forms of MS \cite{scott2020siponimod}. While siponimod has shown promising improvements in clinical outcomes for patients, there is still a need for better binding ligands to fully target S1PR1 and further advance treatment options for MS \cite{synnott2020effectiveness}. By illuminating the conformational properties of ligands that bind to S1PR1, we can develop better binding ligands as potential treatments for MS. 

\textit{De novo} molecular design is a powerful tool for discovering new drug candidates \cite{meyers2021novo}; however, traditional methods can be slow and limited in their creativity. The median estimated research and development cost to bring a new drug to market is around \$1 billion, with a timeline spanning 12 to 15 years. Most of the cost is attributed to the time and resources spent searching for the ideal drug candidate amid a vast chemical space \cite{hughes2011principles}. In recent years, deep generative modeling has emerged as a promising alternative for rapidly screening potential drug candidates \cite{sousa2021generative}. Generative models possess the capability to generate structurally similar variations of target molecules as well as molecules that exhibit comparable chemical properties given a set of training molecules \cite{hirohara2018convolutional}. These models can then be used in conjunction with docking simulations to evaluate the binding affinity and interactions of those generated molecules with the target receptor. The most promising candidates are then selected for further testing and development as potential drug therapies \cite{jeon2020autonomous}.

Generative models have been applied in a variety of studies to generate novel ligands with high docking scores and other desirable properties. For example, Im et al. generated single-strand nucleic acid sequences that bind to a target protein with high affinity and specificity \cite{im2019generative}. Jeon et al. used a deep generative model to find potent novel agonists for the D4 dopamine receptor that have high docking scores \cite{jeon2020autonomous}. Additionally, there have been many studies in recent years using generative modeling to identify new potential COVID-19 drug candidates \cite{chenthamarakshan2020cogmol, amilpur2022predicting}. Generative models have also been widely used to optimize other drug-relevant properties such as the partition coefficient and aqueous solubility \cite{maziarka2020mol, bilodeau2022generating}. The success of these studies prompted us to examine how deep generative models could be used for finding potential ligand candidates for S1PR1.

\begin{figure}[htb]
  \centering
  \includegraphics[scale = 0.63]{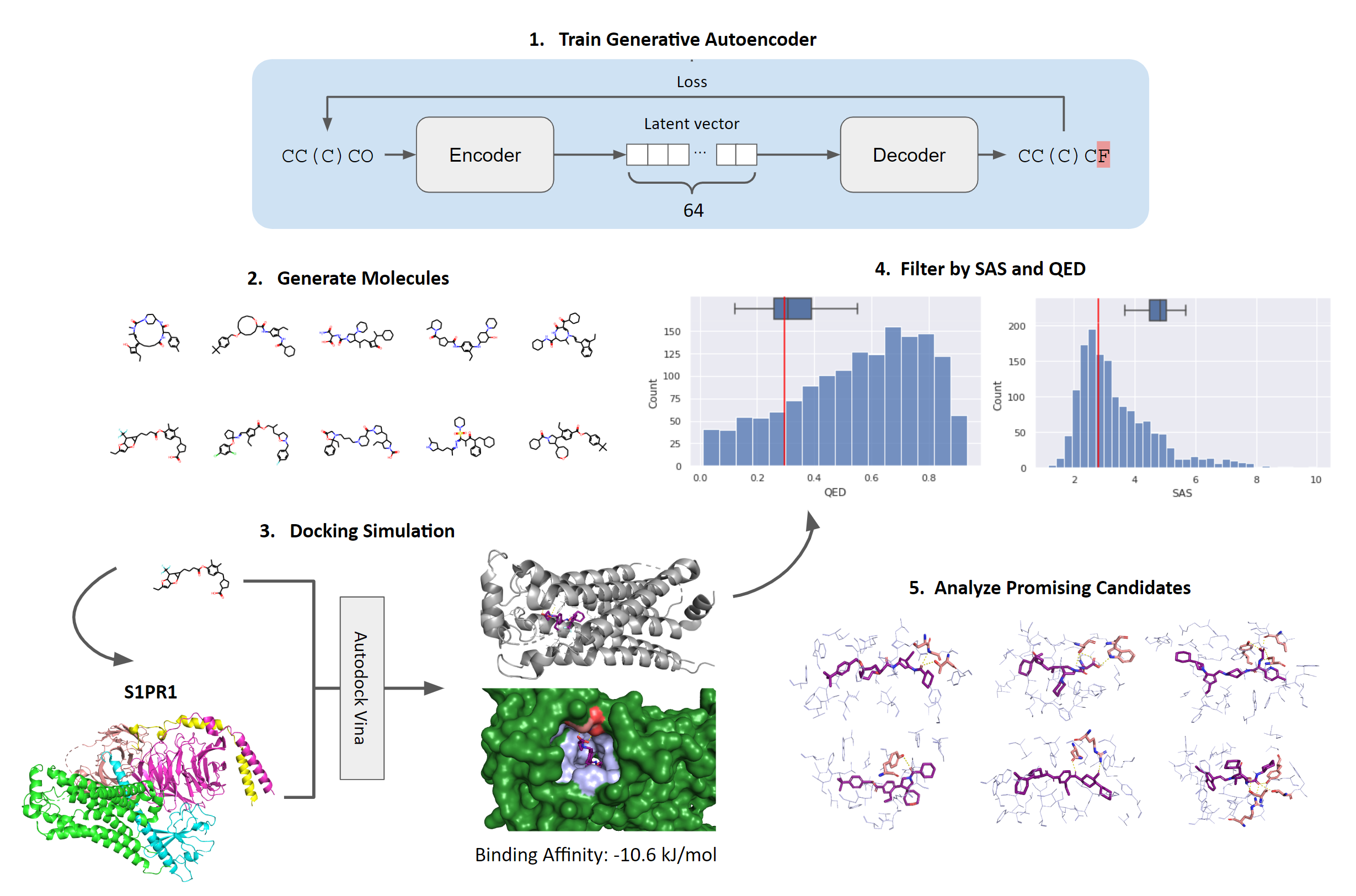}
  \caption{\textbf{Graphical Abstract.}}
  \label{fig:graphical-abstract}
\end{figure}

The primary purpose of this study is to investigate what conformational properties of a ligand contribute to high binding affinity for S1PR1. An analysis of the binding site of S1PR1 using PyMOL \cite{PyMOL} shows that there are multiple hydrophobic residues and a few hydrogen bonding residues on one side within the binding pocket of the receptor. We thus hypothesized that ligands with large nonpolar surface area and a few polar functional groups may bind more effectively to S1PR1. To test this hypothesis, we finetuned a previously characterized machine learning (ML) model to generate a set of candidate ligands and then docked them to S1PR1 using Autodock Vina \cite{trott2010autodock} to evaluate their associated binding affinities \cite{cz4008}. We evaluated these ligands using metrics of druglikeness such as Lipinski’s Rule of Five \cite{lipinski2001lombardo} (\Cref{tab:table1}) and their ease of synthesis. The protein-ligand interactions were further analyzed using PyMOL to understand the mechanisms behind the observed binding affinities (\Cref{fig:graphical-abstract}).

\begin{table}[htb]
 \caption{Lipinski’s Rule of Five}
  \centering
  \begin{tabular}{lll}
    \cmidrule(r){1-2}
    Rule     & Description \\
    \midrule
    MW & The molecular weight should be less than or equal to 500 daltons. \\
    LP & The partition coefficient should be less than or equal to 5. \\
    HBD & There should be no more than five hydrogen bond donors. \\
    HBA & There should be no more than ten hydrogen bond acceptors. \\
    \bottomrule
  \end{tabular}
  \label{tab:table1}
\end{table}

\section{Results}
\label{sec:results}
\subsection{Binding Pocket of S1PR1}

\begin{figure}[h!]
  \centering
  \includegraphics[scale = 0.7]{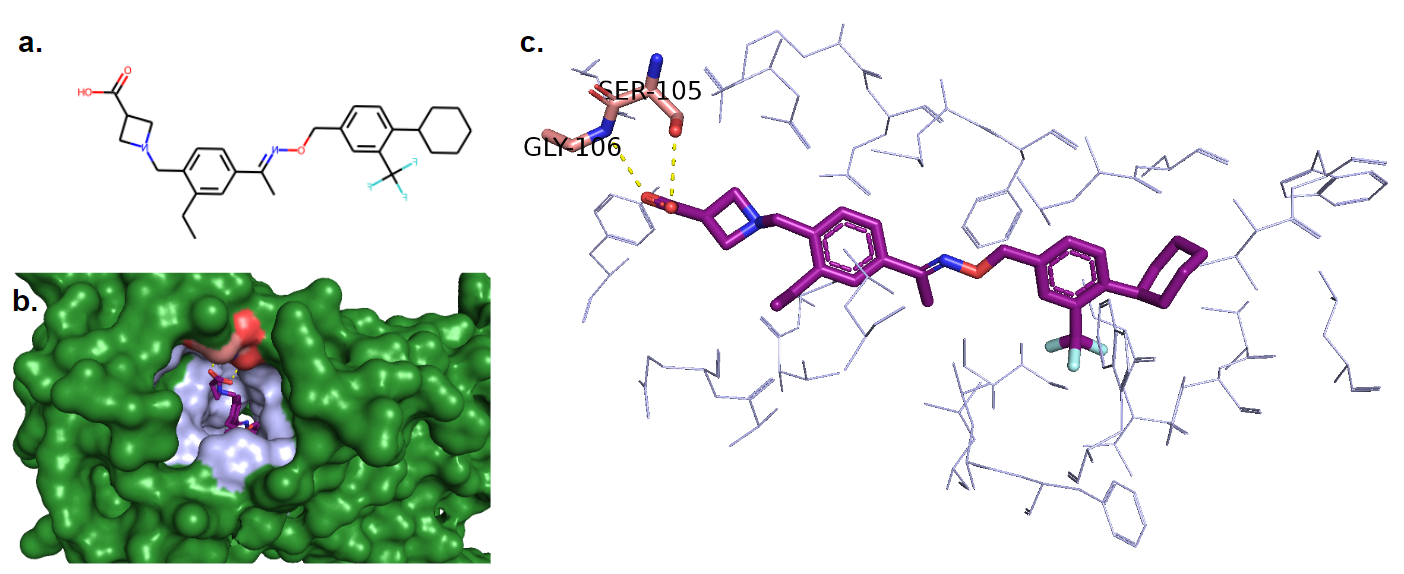}
  \caption{\textbf{Interacting residues with siponimod in the binding pocket of the S1PR1.} A) Structure of siponimod. Note the presence of aromatic rings in the molecule as well as a carboxylic acid. B) Surface-level view of the S1PR1-siponimod complex, with polar-interacting regions (red), other interacting regions (lavender) in the binding site, and the rest of the protein (green). C) Zoomed in view of B highlighting polar interactions by Serine-105 and Glycine-106.}
  \label{fig:fig2}
\end{figure}

We examined the binding pocket of S1PR1 to find the residues that interact with siponimod. The structure of siponimod contains one carboxylic acid on one end and two aromatic rings as well as a trifluoromethyl group, demonstrating the capability to form both polar and nonpolar interactions (\Cref{fig:fig2}a). The molecular weight of siponimod is 516.6 daltons, which is relatively high for drug-like molecules \cite{lipinski2001lombardo}. A surface-level view of the S1PR1-siponimod complex shows the polar and nonpolar regions that interact with siponimod in the binding pocket (\Cref{fig:fig2}b). A closer look at the residues surrounding siponimod reveals two primary residues interacting with siponimod through polar interactions or hydrogen bonds: Glycine-106 and Serine-105. While nonpolar interactions play major roles in the rest of the binding pocket, polar interactions are critical for achieving high binding affinity (\Cref{fig:fig3}c).

\subsection{Model Finetuning and Ligand Generation}

We previously developed and characterized an autoencoder model for generating small molecules with similar properties, trained on a subset of the ChEMBL22 database. Briefly, the model is trained to encode each molecule as a 64-dimensional latent vector and decode the vector back into the original molecule, thus learning the chemically plausible latent space. Novel molecules may be generated by prompting the model with a new latent vector (a “seed”). However, it was found that the model tended to generate molecules similar to its training set \cite{cz4008}.We aimed to leverage this characteristic to generate siponimod-like variants by finetuning the model on a siponimod-like subset of its training set. By mutating the siponimod latent vector, we could generate a variety of latent seeds, with each seed yielding up to several candidate molecules, as the decoding process is probabilistic.

\begin{figure}[htb]
  \centering
  \includegraphics[scale = 0.8]{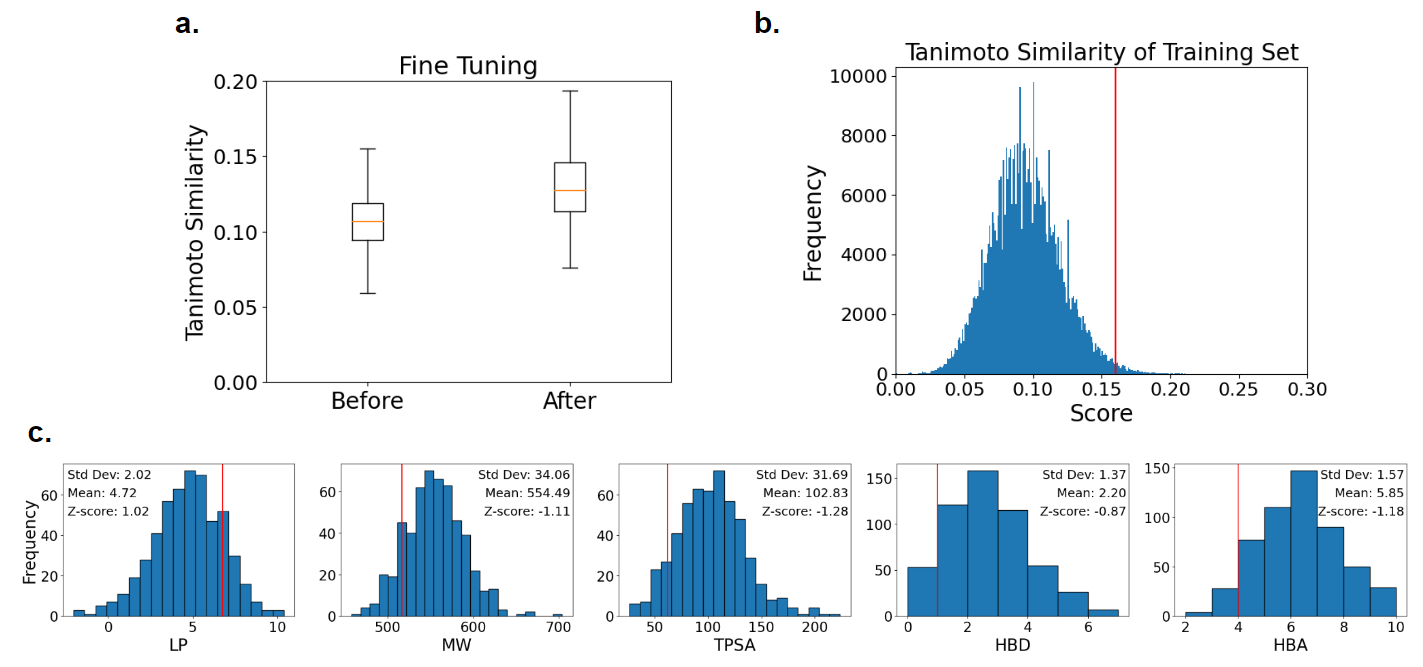}
  \caption{\textbf{Comparing generated ligands with siponimod.} A) Tanimoto similarities of generated ligands with siponimod before and after fine tuning. B) Tanimoto similarities of the 500,000-molecule training set with respect to siponimod. The red line indicates the 99th percentile of Tanimoto similarity (0.16), and the 4505 ligands above this similarity threshold were used to finetune the model. C) Distribution of chemical properties (LP, MW, TPSA, HBD, HBA) for post-finetuning generated ligands, with means and standard deviations shown. The red line indicates the corresponding value for siponimod. For all five properties, the red line lies less than 1.3 standard deviations (z-score) away from the mean.
}
  \label{fig:fig3}
  
\end{figure}

Prior to finetuning, 462 molecules were generated from siponimod-based seeds and it was found that the median Tanimoto similarity score between the majority of the generated ligands and siponimod was 0.107 (\Cref{fig:fig3}a). Using the 99th Tanimoto percentile (corresponding to a similarity score of 0.16) as a cutoff, we compiled a finetuning dataset of 4505 molecules from the original training set with the highest similarity to siponimod (\Cref{fig:fig3}b). After finetuning the model for 50 epochs, we then generated 535 molecules using siponimod as a seed and found a 19\% increase in similarity scores (new median: 0.127) (\Cref{fig:fig3}a). We further analyzed the chemical properties of the post finetuning generated molecules in comparison to the respective properties of siponimod (\Cref{fig:fig3}c). For all five properties examined—the partition coefficient (LP), average molecular weight (MW), topological polar surface area (TPSA), number of hydrogen bond donors (HBD), and number of hydrogen bond acceptors (HBA)—the siponimod baseline lies less than 1.3 standard deviations away from the corresponding mean for the generated molecules, demonstrating that the model recapitulates chemical similarity to some degree.

\subsection{Characteristics of High-Affinity Ligands}

\begin{figure}[H]
  \centering
  \includegraphics[scale = 0.99]{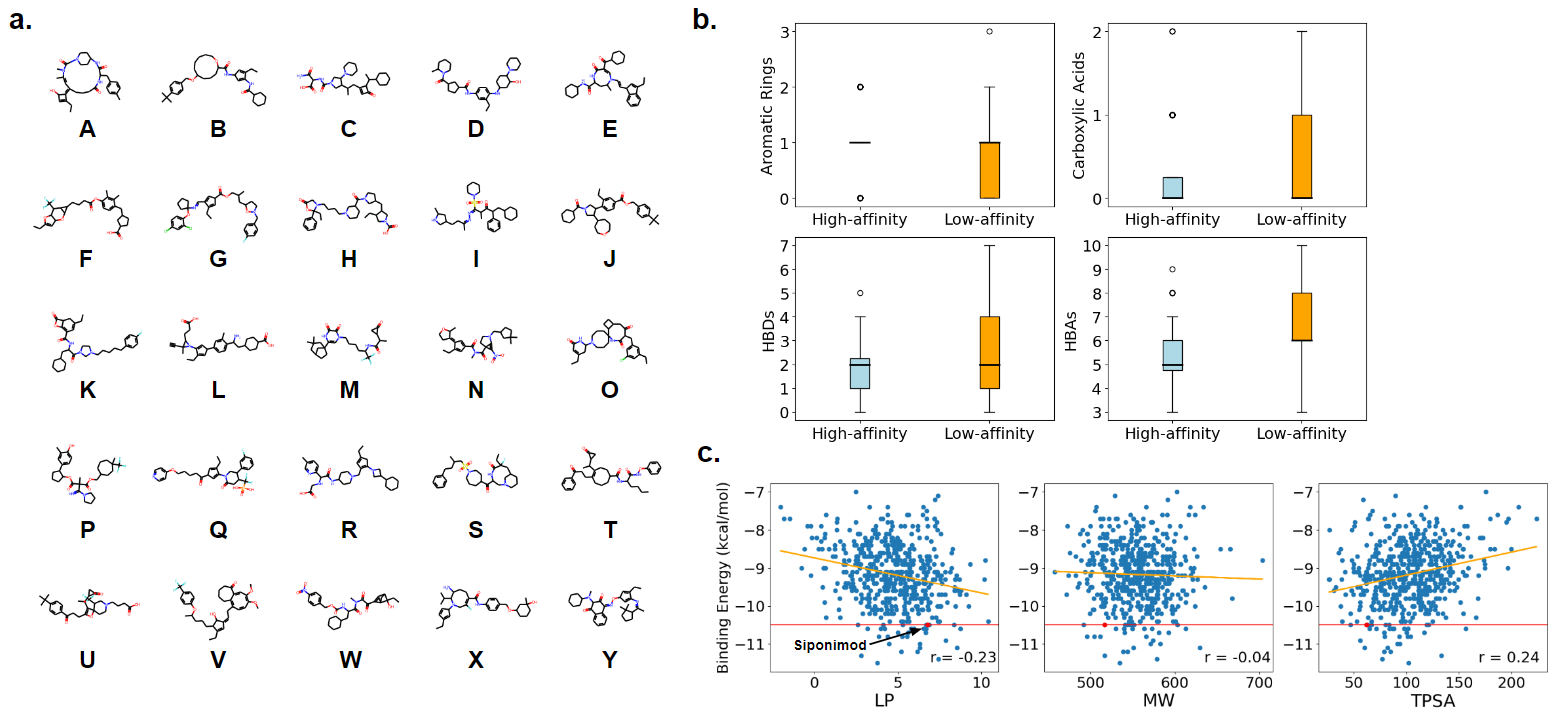}
  \caption{\textbf{High-affinity and low-affinity generated ligands.} A) The 25 highest-affinity generated ligands, all with binding energy $\leq$ -10.6 kcal/mol (the binding energy of siponimod). B) Number of aromatic rings, carboxylic acids, hydrogen bond donors, and hydrogen bond acceptors for the top 100 highest affinity ligands (blue) versus the lowest affinity 100 ligands (orange). C) Correlation between binding energy and chemical properties of interest (partition coefficient, molecular weight, topological polar surface area). A non-negligible trend (indicated by r-value) exists for LP and TPSA, but not molecular weight.}
  \label{fig:fig4}
\end{figure}

We used Autodock Vina to calculate the binding affinity of the post-finetuning generated molecules. Each candidate ligand was repeatedly virtually docked to the binding site, and the conformation with the lowest estimated binding energy was chosen as the most likely conformation. We identified 25 ligands possessing a more favorable binding energy than that of siponimod (-10.6 kcal/mol) (\Cref{fig:fig4}a). As a quantitative metric, we calculated the number of aromatic rings, carboxylic acids, hydrogen bond donors, and hydrogen bond acceptors for the highest affinity 100 ligands versus the lowest affinity 100 ligands (\Cref{fig:fig4}b). The high-affinity ligands tended to have more aromatic rings, but fewer carboxylic acids, HBDs, and HBAs than the low-affinity ligands. We also investigated the correlations between binding affinity and chemical properties (LP, MW, TPSA, HBD, HBA) of generated ligands (\Cref{fig:fig4}c). Higher affinity ligands (lower binding energy) tended to possess higher LP, lower TPSA, fewer HBDs, and fewer HBAs, although these trends were very weak. There was no noticeable correlation between binding affinity and molecular weight.

\subsection{Druglikeness and Synthetic Practicality}

\begin{figure}[H]
  \centering
  \includegraphics[scale = 1.0]{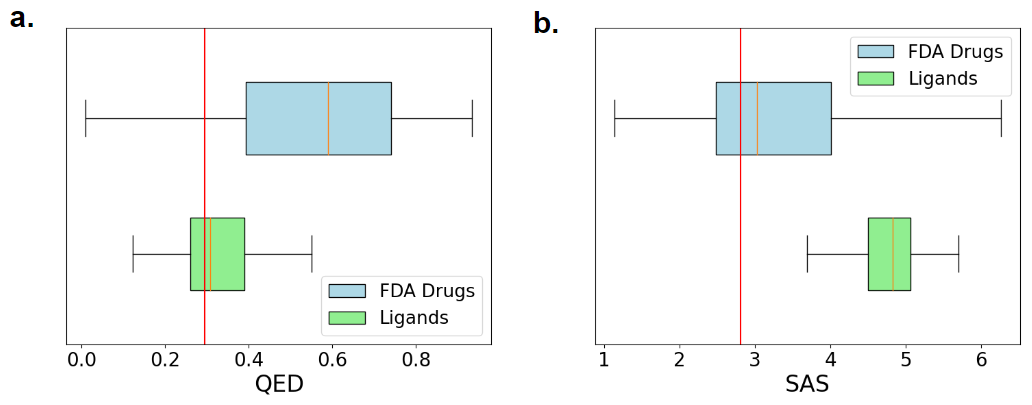}
  \caption{\textbf{Druglikeness and synthetic practicality of high-affinity ligands.} A) Boxplots: Distribution of QED for 2013 FDA approved drugs and distribution of the quantitative estimate of druglikeness (QED) for the 25 high-affinity ligands. Higher is better. Red line is siponimod. B) Boxplots: Distribution of SAS scores for 2013 FDA approved drugs and distribution of the synthetic accessibility score (SAS) for the 25 high-affinity ligands. Lower is better. Red line is siponimod.
}
  \label{fig:fig5}
\end{figure}

High binding affinity is only one of many factors to consider when searching for drug candidates. Two other factors include druglikeness (possessing drug-like properties such as ease of absorption) and synthetic accessibility (ease of synthesis). We compared the quantitative estimate of druglikeness (QED) and synthetic accessibility scores (SAS) for the 25 high-affinity ligands with the QED and SAS scores of all FDA-approved drugs in 2013 \cite{ericminikel} (\Cref{fig:fig5}a, \Cref{fig:fig5}b). Higher QED and lower SAS scores are more desirable. Lipinski’s Rule of Five, another metric for druglikeness consisting of 4 simple rules, was tested for each of the 25 high-affinity ligands as well as siponimod itself (\Cref{tab:table2}). Though most of the generated ligands violate two or more rules, siponimod itself also violates two rules.

\begin{table}[H]
 \caption{\textbf{Lipinski’s Rule of Five is tested for each high-affinity ligand and siponimod.} Green cells indicate obeyed rules. Ligand labels in the first column correspond to those in Figure 4a and Figure 6c. The last row is siponimod. Six ligands (highlighted in orange) are in both the top 50th percentile of QED and bottom 50th percentile of SAS, further visualized in \Cref{fig:fig6}c.}
  \centering
  \includegraphics[scale = 1.5]{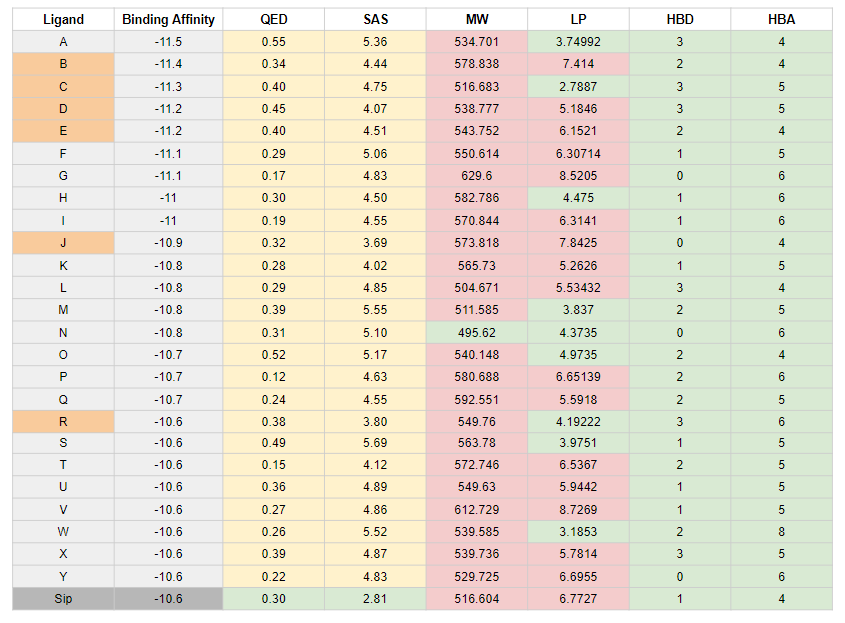}
  \label{tab:table2}
\end{table}

\subsection{Important Binding Residues}

\begin{figure}[H]
  \centering
  \includegraphics[scale = 1.0]{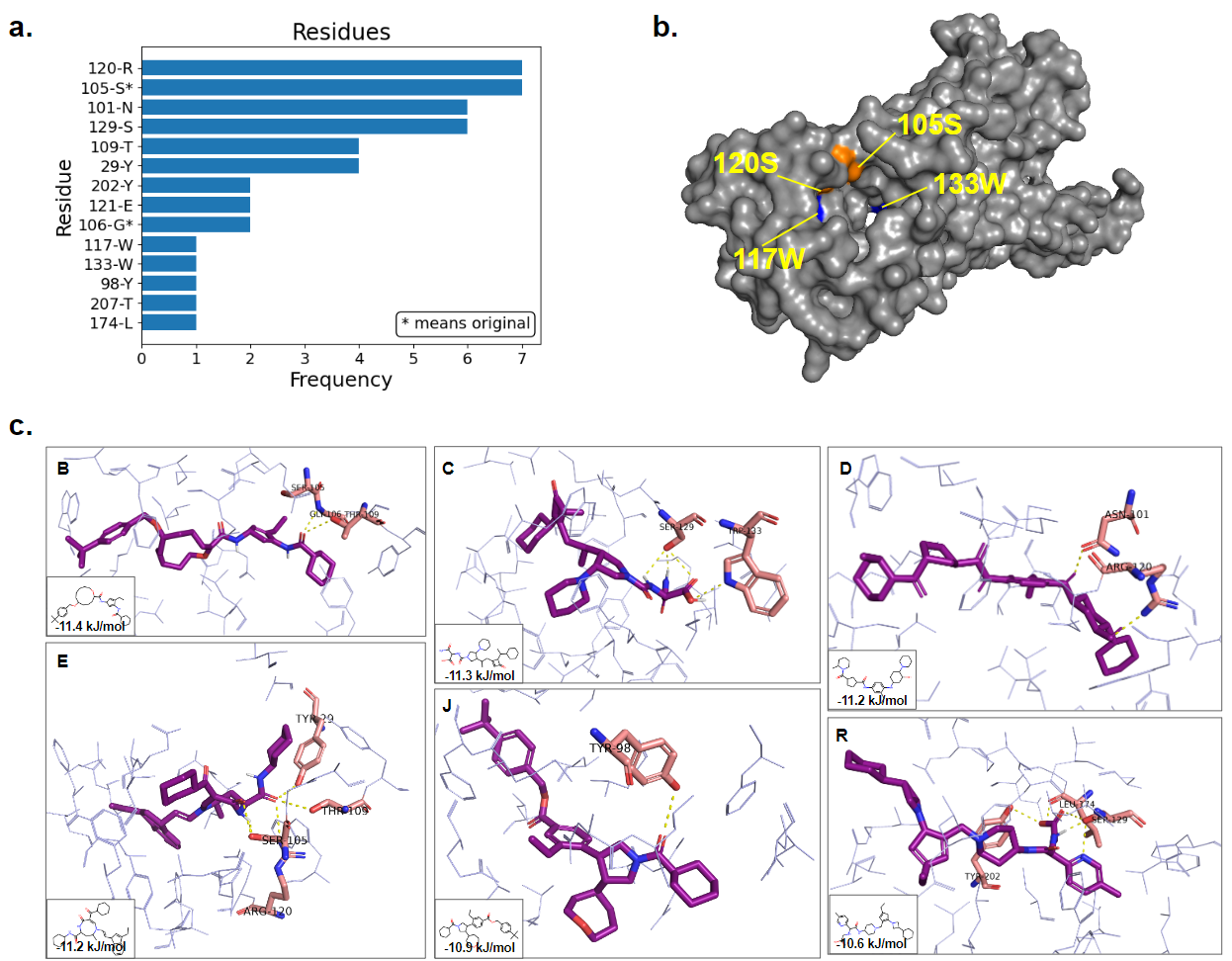}
  \caption{\textbf{Interactions of highest-affinity ligands with binding site.} A) Frequency of interactions between S1PR1 binding pocket residues and the 25 highest-affinity ligands. Only showing residues with at least one interaction. (*) denotes residues implicated in original siponimod-S1PR1 complex. B) S1PR1 space-filling model with the two most frequently polar-interacting residues (120-R and 105-S, orange) and two least frequently polar-interacting residues (117-W and 133-W, blue) highlighted in the binding pocket. C) S1PR1 binding interactions with six ligands (purple) in the top 50th percentile of QED and bottom 50th percentile of SAS. Polar-interacting residues are highlighted in orange. The bottom-left insets show the structure of each ligand along with its binding affinity.
}
  \label{fig:fig6}
\end{figure}

For each of the most promising ligands, we aimed to assess its polar binding interactions with S1PR1. Polar interactions were approximated from the docking simulations by considering all protein-ligand contacts within a threshold of 5 angstroms. We counted the number of times each polar-interacting residue was involved in an interaction with one of the 25 highest-affinity ligands (\Cref{fig:fig6}a). Residues 105-S, 120-R, 129-S, 101-N were most likely to be involved in polar interactions, with 105-S and 106-G also involved in the binding of siponimod. The two most commonly involved residues are near the edge of the binding pocket, while the two least commonly involved residues reside deeper within the binding pocket (\Cref{fig:fig6}b). Finally, we selected the six ligands which were in the top 50th percentile of QED and bottom 50th percentile of SAS and visualized their interactions with the S1PR1 binding site (\Cref{fig:fig6}c). Although the majority of the binding pocket has nonpolar interactions, it is crucial to analyze the residues involved in polar interactions due to their considerable influence on binding affinity.

\section{Discussion}
\label{sec:discussion}
\subsection{Potential of Generative Models in Drug Discovery}
In this study, we assessed the characteristics of high-affinity ligands for the sphingosine 1-phosphate receptor. Based on a qualitative examination of the binding site, we hypothesized that ligands with large nonpolar surface area and one or more polar functional groups will bind more effectively. Candidate ligands were generated with a finetuned SMILES-based autoencoder, using siponimod as a template. The finetuning process was successful in shifting the chemical distribution learned by the generative model towards siponimod-like structures. Surprisingly, nearly 5\% of generated molecules possessed higher simulated binding affinities than siponimod, suggesting that achieving a high theoretical binding affinity is relatively straightforward when other factors such as druglikeness and synthetic accessibility are ignored. These results highlight the potential of ML as a potent tool for drug discovery and the effectiveness of ML-based approaches in optimizing molecular properties for target binding. Moreover, although the present study used an autoencoder-based generative model, more diverse candidate molecules might be achieved by experimenting with other machine learning architectures. 

\subsection{Role of Polar and Nonpolar Interactions in S1PR1}
Consistent with our hypothesis, high-affinity ligands tended to exhibit fewer polar functional groups and more nonpolar aromatic rings compared with low-affinity ligands. These results are consistent with previous studies which experimentally demonstrated the crucial role of hydrophobic interactions between S1PR1 and several of its known ligands \cite{yuan2021structures, yu2022structural}. The structural findings presented here, in combination with previous studies, suggest that the optimal ligand consists of a chain of nonpolar rings terminating in one or more polar functional groups.
The binding site of S1PR1 is composed of several amino acid residues that play a crucial role in ligand binding. In general, polar-interacting residues are located at the edge of the binding site, while hydrophobic residues are found deeper inside. Our high-affinity ligands revealed an array of binding site residues capable of forming polar interactions, many of which are not implicated in the siponimod-S1PR1 binding complex. Our results suggest that residues 105-S and 120-R, which are both shallowly located within the binding region, are the most easily accessible residues in the binding site that allow for polar interactions. We also observed residues in the binding site, such as 117W and 133W, that are infrequently involved in polar interactions, possibly because they are less accessible due to being located deeper inside the binding pocket. Nevertheless, the residues highlighted by our generated high-affinity ligands may help inform future efforts in designing S1PR1 ligands.

\subsection{Future Work}
Using computational estimates of druglikeness and synthetic accessibility, we identified six novel ligands as the most promising potential drug candidates targeting S1PR1. They all have SAS scores between 3 and 4, meaning they are fairly easy to synthesize \cite{ertl2009estimation}. These molecules are also reasonably druglike, possessing QED scores within the range of past FDA-approved drugs \cite{bickerton2012quantifying}. To validate these six ligands as feasible drugs, further work is necessary to verify the results of the docking simulations and collect data on other important metrics such as bioavailability and toxicity. Our research highlights the potential of machine learning models not only for generating novel drug candidates, but also enhancing our understanding of protein-ligand binding when used in conjunction with docking simulations.

\section{Methods}
\label{sec:methods}

\subsection{Model}

The generative model was built in Google Colaboratory with Python 3.7 and TensorFlow version 2.8.2, utilizing the Keras API. The dataset used for training and testing was the CHEMBL22 dataset \cite{gaulton2017chembl} downloaded from Kaggle, a community data science platform. Out of 1.5 million small molecule SMILES strings, 500,000 were randomly chosen, with a max length of 100 characters. The Gated Recurrent Unit (GRU) autoencoder architecture and training procedure is extensively discussed in \cite{cz4008}. After the model was trained for 200 epochs on the original dataset, finetuning with a siponimod-like subset of data was performed. This dataset was comprised of SMILES strings in the original training dataset that had a 99th percentile Tanimoto similarity score with siponimod, and the model was further trained on this new dataset for another 50 epochs. The resulting model was used to generate novel molecules from a siponimod template with 1000 attempts for each seed, resulting in 535 generated ligands.

\subsection{Autodock Vina}
A Python script was used to dock each generated ligand to S1PR1 in Autodock Vina. The protein-ligand complex was first retrieved as a PDB file through the biopython library, using the PDB ID of siponimod (7EO4). S1PR1 was isolated as a separate PDB file in preparation for docking. Using MGLtools, the protein was parameterized to ensure that its physical and chemical properties are correctly represented in the docking simulation. Each ligand was prepared by first converting it from SMILES to PDB format, then parameterizing and adding Gasteiger charges. Gasteiger charges were added to improve accuracy by accounting the effects of electrostatic potential around the protein. Energy minimization was performed for the ligands using a General AMBER Force Field (GAFF) forcefield energy minimization with 10,000 steps.

A gridbox was defined in S1PR1 to specify the location of its binding pocket using the py3dmol library. The ligands were docked using Autodock Vina using the default number of docking poses (9). The highest binding affinity (lowest binding energy) for each ligand was saved to a dictionary, and its associated output file (specifying the position) was also downloaded. The final dictionary was filtered for ligands with binding energies lower than or equal to -10.6 kcal/mol (the binding energy of siponimod), resulting in a total of 25 “high-affinity” ligands.

\subsection{Protein-Ligand Interactions}
To analyze protein-ligand interactions, PyMOL was used to open the cleaned protein structure along with the optimal binding pose of each “high-affinity” ligand. The polar-interacting residues were identified using "Find Polar Contacts" in PyMOL.

\subsection{Plots}
RDKit 2021.09.4 was used to estimate the chemical properties and other metrics of the ligands under investigation, including molecular weight, logP, number of hydrogen bond acceptors and donors, and QED scores. The library was also used to display the molecular structures of the ligands. SAS scores were calculated using a Python script by Ertl and Shuffenhauer \cite{ertl2009estimation}. Matplotlib and Seaborn, two popular Python libraries for data visualization, were used to create plots such as histograms, box plots, and scatter plots.

\clearpage
\bibliographystyle{unsrt}  
\bibliography{references}

\end{document}